\documentclass[11pt]{article}

\usepackage{acl}
\usepackage{booktabs}
\usepackage{times}
\usepackage{latexsym}
\usepackage{amsfonts} 
\usepackage{tikz}
\usetikzlibrary{arrows.meta, positioning}

\usepackage[T1]{fontenc}

\usepackage[utf8]{inputenc}

\usepackage{microtype}

\usepackage{inconsolata}
\usepackage{amsmath}
\usepackage{graphicx}

\title{An Encoder-Integrated PhoBERT with Graph Attention for Vietnamese Token-Level Classification}

\author{Ba-Quang Nguyen \\
  University of Engineering and Technology \\
  Vietnam National University \\
  Hanoi, Vietnam \\
  \texttt{23020412@vnu.edu.vn}
}

\usepackage{float}
\begin{document}
\maketitle
\begin{abstract}
In this paper, we propose a novel neural architecture named \textbf{TextGraphFuseGAT}, which integrates a pretrained transformer encoder (PhoBERT) with Graph Attention Networks (GAT) for token-level classification tasks. The proposed model constructs a fully connected graph over the token embeddings produced by PhoBERT, enabling the application of a GAT layer to capture rich inter-token dependencies beyond those modeled by sequential context alone. To further enhance contextualization, a Transformer-style self-attention layer is applied on top of the graph-enhanced embeddings. The final token representations are passed through a classification head to perform sequence labeling.

We evaluate the effectiveness of our approach on three Vietnamese benchmark datasets: the \textbf{PhoNER-COVID19} dataset \cite{truong2021covid} for Named Entity Recognition (NER) in the COVID-19 domain, the \textbf{PhoDisfluency} dataset \cite{dao2022disfluency} for speech disfluency detection, and the \textbf{VietMed-NER} dataset \cite{le2024medical} for medical-domain NER. VietMed-NER is the first Vietnamese medical spoken NER dataset, featuring 18 entity collected from real-world medical speech transcripts, annotated using the standard BIO tagging scheme. Its specialized vocabulary and complex domain-specific expressions make it a challenging benchmark for token-level classification models.

Experimental results demonstrate that our method consistently outperforms strong baselines, including transformer-only and conventional neural models such as BiLSTM + CNN + CRF, confirming the effectiveness of combining pre-trained semantic features with graph-based relational modeling for improved token classification across multiple domains.
\end{abstract}

\section{Introduction}

\subsection{Background and Motivation}
Token-level classification is a fundamental task in NLP, supporting applications such as Named Entity Recognition (NER), POS tagging, and disfluency detection, which in turn enable downstream systems like information extraction and dialogue processing. For Vietnamese, a morphologically rich language with limited annotated resources, accurate token-level modeling is especially important.

Pre-trained language models such as PhoBERT have substantially advanced Vietnamese NLP by providing strong contextual embeddings. However, self-attention alone may be insufficient to capture explicit syntactic structures or long-range discourse cues, motivating hybrid architectures that combine contextual encoders with structural modeling.

\subsection{Challenges in Vietnamese Token-Level Tasks}
Vietnamese token-level classification remains difficult due to several factors:  
(i) \textbf{Ambiguous tokenization}, since whitespace does not always align with word boundaries;  
(ii) \textbf{Data scarcity}, with limited and domain-specific annotated corpora;  
(iii) \textbf{Severe label imbalance}, where the majority “O” class dominates and rare entities are underrepresented.  
In addition, each task poses unique difficulties: NER requires distinguishing overlapping entity types, while disfluency detection relies on subtle discourse patterns in the absence of prosodic features.
\subsection{Datasets}

We evaluate our approach on three publicly available Vietnamese benchmarks: \textbf{PhoNER-COVID19}, \textbf{PhoDisfluency}, and \textbf{VietMed-NER}. These datasets span two distinct domains of Vietnamese language processing: formal written text and spontaneous speech transcripts.

\textbf{PhoNER-COVID19} is a Vietnamese NER dataset focusing on short sentences related to the COVID-19 pandemic. It contains 10 entity types (e.g., \textit{PATIENT\_ID}, \textit{SYMPTOM\&DISEASE}, \textit{LOCATION}, \textit{DATE}) annotated with the standard BIO scheme, yielding 20 distinct labels. The data is sourced from news headlines, official government announcements, and social media posts, capturing a realistic snapshot of Vietnamese usage during the pandemic. Both word-level and syllable-level tokenizations are provided. The training/validation/test splits contain 5,027/2,000/3,000 sentences, corresponding to 132,511/56,283/85,678 tokens for word-level and 167,541/71,325/108,354 tokens for syllable-level. No nested entities are annotated.

\textbf{PhoDisfluency} is a Vietnamese disfluency detection dataset annotated at both word and syllable levels. It follows the disfluency structure of Shriberg (1994), with three label categories: \textit{RM} (Reparandum), \textit{IM} (Interregnum), and \textit{O} (fluent words). The training/validation/test sets comprise 4,478/500/893 sentences, corresponding to 82,419/10,238/16,751 tokens for word-level and 98,694/12,024/19,850 tokens for syllable-level. The label set is consistent across all splits, and no nested annotations are included.

\textbf{VietMed-NER} is a Vietnamese medical NER dataset manually annotated with 18 entity types (e.g., \textit{DISEASESYMPTOM}, \textit{DRUGCHEMICAL}, \textit{SURGERY}) using the BIO scheme. The corpus is collected from ASR transcripts of Vietnamese medical speech, covering terminology from various specialties. It contains 9,270 sentences. The training/validation/test splits include 4,620/1,150/3,500 sentences with 11,109/2,729/7,897 entity mentions, respectively. The dataset is provided at the word level without nested entities.

Overall, these datasets provide a diverse and challenging evaluation benchmark for token-level classification in both formal written news and informal spoken language domains. We use both word-level and syllable-level versions (where available) to assess the generalizability and robustness of our method across different linguistic granularities.

\subsection{Contributions}

Our main contributions in this work are:
\begin{itemize}
    \item We introduce \textbf{TextGraphFuseGAT}, a hybrid model that combines transformer-based contextual embeddings with graph attention mechanisms for token-level classification.
    \item We evaluate our model on three challenging Vietnamese benchmarks, demonstrating superior performance compared to transformer-only baselines.
    \item We provide empirical evidence that relational modeling via graphs complements sequential modeling, suggesting a promising direction for hybrid neural architectures, especially in under-resourced languages like Vietnamese.
    \item To our knowledge, this is the first work to integrate graph attention and transformer decoding into a unified architecture for Vietnamese token-level tasks.
\end{itemize}
\section{Related Work}
Token-level classification tasks, such as disfluency detection and Named Entity Recognition (NER), are central challenges in natural language processing (NLP). For Vietnamese, both tasks have received increased attention in recent years due to the need for high-quality language technology resources in under-resourced languages.
\subsection{Disfluency Detection}
Disfluency detection has been studied extensively in English. Early neural approaches such as Zayats et al. (2016) \cite{zayats2016disfluency} proposed a BiLSTM model that significantly outperformed traditional pattern-based methods. Later, prosodic and attention-based signals were integrated into disfluency modeling (Zayats and Ostendorf, 2019 \cite{zayats2019giving}). Schlangen and Hough (2017) \cite{schlangen2017joint} jointly modeled disfluency detection and utterance segmentation from speech, highlighting the close interaction between the two problems. Wang et al. (2020) \cite{wang2020combining} explored self-training and self-supervised learning for unsupervised disfluency detection, while Lou and Johnson (2020) \cite{jamshid-lou-johnson-2020-improving} introduced semi-supervised self-training with self-attentive parsing, setting new benchmarks on English datasets.

For Vietnamese, Dao et al. (2022) \cite{dao2022disfluency} presented one of the first comprehensive studies targeting Vietnamese disfluencies. They explored various architectures, including BiLSTM + CNN + CRF, XLM-R, and PhoBERT fine-tuning. Their experiments showed that pretrained language models like PhoBERT significantly outperformed traditional sequence labeling architectures.

\subsection{Named Entity Recognition}
Named Entity Recognition (NER) has long been a fundamental task in NLP, aiming to identify and classify proper names such as persons, organizations, locations, and other domain-specific entities. In English, large-scale shared benchmarks such as CoNLL-2003 \cite{sang2003introduction} has served as standard testbeds for decades. Early approaches relied on hand-crafted features combined with sequence labeling models such as Conditional Random Fields (CRFs) \cite{lafferty2001conditional} and BiLSTM-CRF architectures \cite{lample2016neural}, which already demonstrated strong performance by capturing both lexical and contextual cues. 

With the introduction of pretrained contextualized embeddings, particularly ELMo \cite{peters-etal-2018-deep} and later transformer-based encoders such as BERT \cite{devlin2019bert} and RoBERTa \cite{liu2019roberta}, the state of the art on English NER benchmarks was dramatically advanced. These models leverage large-scale pretraining to capture both syntactic and semantic information, leading to substantial improvements over purely supervised methods. More recent work has further explored domain adaptation and multilingual modeling, where cross-lingual encoders such as XLM-R \cite{conneau2019unsupervised} have been shown to transfer NER capabilities across languages, including low-resource settings.

NER has also been widely studied for Vietnamese. The VLSP shared task \cite{nguyen2018vlsp} provided an early benchmark dataset and evaluation campaigns. Vu et al. (2018) introduced VnCoreNLP \cite{vu2018vncorenlp}, a widely used toolkit for multiple Vietnamese NLP tasks, including NER. Truong et al. (2021) \cite{truong2021covid} introduced a domain-specific COVID-19 dataset and benchmarked several models, again confirming that PhoBERT fine-tuning achieved superior results. More recently, Le-Duc et al. (2024) \cite{le2024medical} proposed \textbf{VietMed-NER}, a large-scale Vietnamese medical NER dataset containing 18 entity types collected from medical records, clinical notes, and other healthcare-related documents.

\subsection{Graph-based Models for Token Classification}
Recent advances in graph neural networks (GNNs) have demonstrated their effectiveness for modeling linguistic structures beyond sequential order. Marcheggiani and Titov (2017) \cite{marcheggiani2017encoding} applied graph convolutional networks (GCNs) to semantic role labeling, while Cetoli et al. (2017) \cite{cetoli2017graph} used GCNs for named entity recognition. Zhang et al. (2018) \cite{zhang2018graph} showed that pruning dependency trees before applying GCNs improved relation extraction. Velickovic et al. (2018) \cite{velivckovic2018graph} introduced the Graph Attention Network (GAT), which enables more flexible relational modeling by attending over neighborhoods. More recent work such as Zhang et al. (2020) \cite{zhang2020graph} proposed Graph-BERT to unify attention and graph structures.. These studies suggest that relational modeling via graphs can complement sequential modeling in structured prediction tasks.
\subsection{Discussion}
Although these studies highlight the power of pretrained models for Vietnamese, they mostly treat the problem using either sequential models (e.g., BiLSTM-CRF) or transformer-based models in isolation. In contrast, our approach introduces a hybrid graph-based architecture that models token-level relations more explicitly using graph attention mechanisms, followed by a Transformer decoder for compositional refinement — a combination that is underexplored in Vietnamese token classification.

Recent work has shown that graph neural networks (GNNs) are effective in modeling token relations beyond linear order \cite{wu2020comprehensive}, while Transformer decoders have been adapted to structured prediction tasks \cite{zhou2020transformers}. To the best of our knowledge, no prior work has proposed an architecture that jointly integrates these two components within a unified framework for disfluency detection or NER, particularly in the context of low-resource languages.

This motivates our proposed model: a PhoBERT-augmented, graph-enhanced, transformer-decoder-based token classifier for Vietnamese.

Summary of existing approaches:
\begin{itemize}
\item BiLSTM + CNN + CRF architectures for token classification.
\item Pretrained language models (PhoBERT, XLM-R) fine-tuned for Vietnamese.
\item GNNs and Transformers used separately in prior work.
\item No prior work combines GAT + Transformer decoder for Vietnamese token-level tasks.
\end{itemize}

\section{Method}

\subsection{Pretrained Embedding Encoder}
We employ PhoBERT \cite{nguyen2020phobert}, a state-of-the-art Vietnamese 
language model based on RoBERTa, as the backbone encoder in our architecture. 
Given an input sentence, we tokenize it using PhoBERT's Byte-Pair Encoding (BPE) tokenizer 
and extract contextual embeddings from the final hidden layer. 
These token-level representations encode rich syntactic and semantic information. 
Unlike approaches that keep the pre-trained encoder frozen, we fine-tune PhoBERT jointly 
with the downstream components, allowing the contextual representations to adapt dynamically 
to the characteristics of each specific task. 
The PhoBERT embeddings $H \in \mathbb{R}^{n \times d}$ are then used as input node features 
for the subsequent graph construction module. For simplicity, we omit the batch dimension $B$; all representations can be extended to $\mathbb{R}^{B \times n \times d}$ in practice.

\subsection{Graph Construction Module}
To explicitly model interactions among tokens, we construct a fully connected graph 
over each input sequence, treating every token as a node and connecting all token pairs 
via directed edges. Each node is also connected to itself.
The node features are initialized as the PhoBERT embeddings $H$, 
so the output of this module is the graph $(G, H)$, 
which serves as input to the GAT layer. 
We build the edge index efficiently over mini-batches by offsetting token indices per sample.

\subsection{Graph Attention Network (GAT)}
We apply a multi-head Graph Attention Network (GAT) \cite{velivckovic2018graph} to the constructed token graph $(G, H)$. GAT computes attention coefficients between nodes via learned functions over node features, rather than dot products as in Transformer self-attention \cite{velivckovic2018graph}, thereby introducing a complementary inductive bias. Prior work has shown that incorporating graph-based biases can enhance Transformer architectures on structured data \cite{dwivedi2020generalization, chen2022structure}. This enables the model to capture diverse relational patterns beyond those represented by self-attention alone. Moreover, multi-head graph attention facilitates modeling of multiple relation types in parallel \cite{velivckovic2018graph, vaswani2017attention}, while also allowing extensions to sparse or linguistically motivated graphs. We apply dropout to both the normalized attention coefficients and the hidden features to prevent overfitting. The output is a set of graph-augmented token representations $H^{\text{gat}} \in \mathbb{R}^{n \times d}$, which are passed to the Transformer decoder.

\subsection{Transformer Decoder Layer}
Following the GAT module, we apply a Transformer decoder \cite{vaswani2017attention} 
to further refine token representations. 
In our design, both the target ($tgt$) and memory ($memory$) inputs are set to 
the GAT-enhanced embeddings, i.e., $tgt = memory = H^{\text{gat}}$. We adopt the decoder layer rather than an encoder layer to leverage its cross-attention mechanism as an additional refinement step, providing a complementary view beyond standard self-attention.
We adopt the standard \texttt{TransformerDecoderLayer} from PyTorch without 
any modification, and the cross-attention mechanism is preserved. 
Therefore, the decoder does not serve as an autoregressive generator, 
but rather as a refinement layer that re-attends to the same sequence. 
This additional refinement outputs $H^{\text{dec}} \in \mathbb{R}^{n \times d}$, 
which captures richer contextual and compositional dependencies 
while preserving the graph-augmented semantics. 
The refined embeddings $H^{\text{dec}}$ are then passed to the classification layer.

\subsection{Classification Layer}
Refined token embeddings $H^{\text{dec}}$ are passed through a feed-forward 
classification layer to produce token-level label predictions. 
For training, we use the standard \texttt{CrossEntropyLoss} in PyTorch 
with an \texttt{ignore\_index} of $-100$ to mask padding and subword tokens.

\subsection{Training Strategy}
We jointly fine-tune PhoBERT together with the graph-based and classification components, 
using AdamW with moderate learning rates. This joint optimization strategy is designed 
to reduce overfitting and training time, making the model effective for low-resource 
sequence labeling tasks.

Our modular design leverages strong pretrained representations, relational reasoning, and lightweight adaptation layers to yield competitive results in low-resource, token-level sequence labeling tasks.

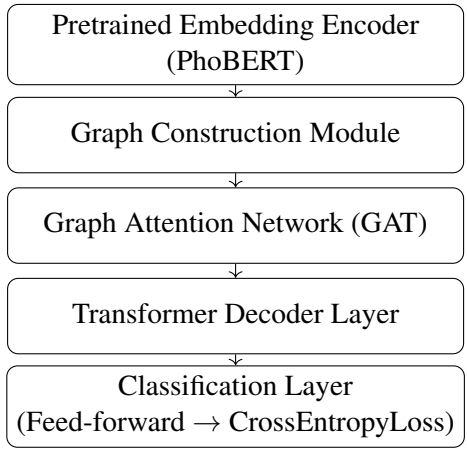
\begin{figure}[h]
\centering
\begin{tikzpicture}[node distance=1.2cm]

\tikzstyle{block} = [rectangle, draw, rounded corners,
                     minimum width=6cm, minimum height=1cm, align=center]

\node[block] (encoder) {Pretrained Embedding Encoder \\ (PhoBERT)};
\node[block, below of=encoder] (graph) {Graph Construction Module};
\node[block, below of=graph] (gat) {Graph Attention Network (GAT)};
\node[block, below of=gat] (decoder) {Transformer Decoder Layer};
\node[block, below of=decoder] (classifier) {Classification Layer \\ (Feed-forward $\rightarrow$ CrossEntropyLoss)};

\draw[->] (encoder) -- (graph);
\draw[->] (graph) -- (gat);
\draw[->] (gat) -- (decoder);
\draw[->] (decoder) -- (classifier);

\end{tikzpicture}
\caption{Overall architecture of the proposed model.}
\end{figure}

\section{Experiments}

In this section, we conduct a comprehensive comparison between our proposed method and prior state-of-the-art approaches on two fundamental Vietnamese token-level classification tasks: Named Entity Recognition (NER) and Disfluency Detection. For the NER task, we evaluate our model on two benchmarks: the COVID-19 NER dataset \cite{truong2021covid} and the VietMed-NER dataset \cite{le2024medical}, benchmarking against strong baselines such as BiLSTM-CRF, XLM-R, and PhoBERT models, with reference results reported by prior work.

Similarly, for the Disfluency Detection task, we adopt the same evaluation setting as \cite{dao2022disfluency}, who showed that fine-tuning pretrained models like PhoBERT and XLM-R yields superior results. By fine-tuning PhoBERT within our architecture and augmenting it with graph-based relational reasoning and transformer-based decoding, we aim to push performance even further.

Our primary goal is to demonstrate that a hybrid architecture, which integrates pretrained embeddings with structural inductive bias (via GAT and Transformer decoder), can effectively boost token-level prediction performance. By jointly optimizing all components—including the pretrained language model—we show that the model not only captures deep contextual semantics but also learns to reason over token interactions, ultimately surpassing or matching strong baselines. This is especially impactful for low-resource languages like Vietnamese, where architectural innovations can complement limited annotated data.

\subsection{Training Settings}

We use PhoBERT\textsubscript{large} as the backbone encoder for both tasks. For Named Entity Recognition (NER), the training is conducted using a maximum sequence length of 128 tokens, batch size of 16, and trained over 15 epochs with a learning rate of 5e-5. Following common practice in Transformer-based models, we apply a weight decay of 0.01 to prevent overfitting, gradient clipping at 1.0 to stabilize training, a warmup ratio of 0.1 to smooth the early optimization phase, and dropout probability of 0.3 to improve generalization. The GAT module is configured with 8 attention heads and a hidden size of 256, which are typical settings ensuring sufficient model capacity without excessive computation. For the PhoNER COVID-19 dataset, we follow \cite{nguyen2020phobert} and use a learning rate of 5e-5, where PhoBERT achieves its best performance. For the more complex VietMed-NER dataset, we use a slightly smaller learning rate of 3e-5.Compared to the original paper, we adopt a much smaller batch size and fewer epochs for each dataset. Specifically, for \textbf{PhoNER}, our setting uses a batch size of 16 and 15 epochs, whereas the original paper used a batch size of 32 and 30 epochs. For \textbf{VietMed-NER}, we also train with a batch size of 16 and 15 epochs, while the original paper used a batch size of 50 and 50 epochs. This adjustment enables faster convergence and reduces computational demand, making the approach feasible on limited resources without compromising performance. Our experiments confirm that the model still converges effectively under these lighter settings, achieving strong performance with only 15 epochs and batch size 16 for PhoNER and VietMed, and 10 epochs for Disfluency Detection, as presented in the \textit{Main Results}. 

For Disfluency Detection, we further adapt the training strategy to reflect the lower complexity of this dataset, which is smaller in size and contains far fewer label types compared to PhoNER and VietMed-NER. Consequently, we train the model for only 10 epochs, which is sufficient for convergence while still optimizing computational efficiency. To match this number of epochs, we experimented with several typical learning rates and selected 2e-5, which achieved the highest performance. In addition, the number of GAT attention heads is reduced from 8 to 4. This reduction lowers computational overhead and prevents over-parameterization, while maintaining adequate modeling capacity for the simpler structure of the disfluency detection task.

The optimizer used in all experiments is AdamW. We train all models using PyTorch and HuggingFace Transformers libraries with early stopping based on validation Micro-F1 score. In terms of efficiency, training on the most complex dataset (VietMed-NER) takes approximately 2 hours on a single NVIDIA T4 GPU (Google Colab environment), showing that the proposed architecture remains computationally feasible even under the heaviest training setting.

\subsection{Evaluation Metrics}

We evaluate both tasks using Micro-F1 and Macro-F1 scores. For NER, we further report per-entity F1 scores to assess model performance across different entity types. For Disfluency Detection, we report F1 scores for RM (repair markers) and IM (interruption markers) categories in addition to the overall Micro-F1.

\subsection{Main Results}

Below, we summarize the comparison results across both tasks, highlighting key performance metrics and gains.

We evaluate our method against strong baselines across three datasets. 
On the PhoNER-COVID19 dataset, we compare BiLSTM-CRF, XLM-R$_{base}$, XLM-R$_{large}$, and our method (Table~\ref{tab:ner_comparison}). 
On the PhoDisfluency dataset, we compare BiLSTM-CRF, XLM-R, PhoBERT, and our method (Table~\ref{tab:word_syllable_comparison}). 
On the VietMed-NER dataset, we report results against BARTpho, mBART-50, PhoBERT, ViDeBERTa, ViT5, XLM-R, and our method 
(Tables~\ref{tab:ner_results} and \ref{tab:vietmed-ner-results}). 
For fair comparison, baseline scores are taken from prior work 
(\cite{truong2021covid} for PhoNER-COVID19; \cite{dao2022disfluency} for PhoDisfluency; \cite{le2024medical} for VietMed-NER).

\begin{table*}[t]
\centering
\resizebox{\textwidth}{!}{%
\begin{tabular}{lcccccccccccc}
\toprule
Model & PAT & PER & AGE & GEN & OCC & LOC & ORG & SYM & TRA & DAT & Mic-F1 & Mac-F1 \\
\midrule
\multicolumn{13}{l}{\textbf{Syllable}} \\
BiL-CRF           & 0.953 & 0.855 & 0.943 & 0.947 & 0.588 & 0.915 & 0.808 & 0.801 & 0.794 & 0.976 & 0.906 & 0.858 \\
XLM-R$_{base}$    & 0.978 & 0.902 & 0.957 & 0.842 & 0.506 & 0.941 & 0.842 & 0.858 & 0.924 & 0.982 & 0.925 & 0.879 \\
XLM-R$_{large}$   & 0.982 & 0.933 & 0.962 & 0.958 & 0.692 & 0.943 & 0.853 & 0.854 & 0.943 & 0.987 & 0.938 & 0.911 \\
\textbf{Our Method} & \textbf{0.990} & \textbf{0.942} & \textbf{0.969} & \textbf{0.976} & \textbf{0.839} & \textbf{0.971} & \textbf{0.923} & \textbf{0.923} & \textbf{0.984} & \textbf{0.991} & \textbf{0.982} & \textbf{0.954} \\
\midrule
\multicolumn{13}{l}{\textbf{Word}} \\
BiL-CRF           & 0.953 & 0.874 & 0.950 & 0.947 & 0.605 & 0.911 & 0.831 & 0.799 & 0.902 & 0.976 & 0.910 & 0.875 \\
PhoBERT$_{base}$  & 0.981 & 0.903 & 0.962 & 0.954 & 0.749 & 0.943 & 0.870 & 0.883 & 0.966 & 0.987 & 0.942 & 0.920 \\
PhoBERT$_{large}$ & 0.980 & 0.944 & 0.967 & 0.968 & 0.791 & 0.940 & 0.876 & 0.885 & 0.967 & 0.989 & 0.945 & 0.931 \\
\textbf{Our Method} & \textbf{0.987} & \textbf{0.961} & \textbf{0.978} & \textbf{0.985} & \textbf{0.841} & \textbf{0.969} & \textbf{0.924} & \textbf{0.928} & \textbf{0.983} & \textbf{0.990} & \textbf{0.984} & \textbf{0.958} \\

\bottomrule
\end{tabular}}
\caption{NER performance on the PhoNER-COVID19 dataset.}
\label{tab:ner_comparison}
\end{table*}
\begin{table}[t]
\centering
\begin{tabular}{lccc}
\toprule
Model & RM-F1 & IM-F1 & Mic-F1 \\
\midrule
\multicolumn{4}{l}{\textbf{Syllable}} \\
BiL-CRF         & 0.882 & 0.947 & 0.915 \\
XLM-R$_{base}$  & 0.946 & 0.977 & 0.962 \\
XLM-R$_{large}$ & 0.953 & 0.978 & 0.966 \\
\textbf{Our Method} & \textbf{0.978} & 0.992 & 0.993 \\
\midrule
\multicolumn{4}{l}{\textbf{Word}} \\
BiL-CRF          & 0.894 & 0.946 & 0.921 \\
PhoBERT$_{base}$ & 0.956 & 0.972 & 0.965 \\
PhoBERT$_{large}$ & 0.953 & 0.981 & 0.968 \\
\textbf{Our Method} & \textbf{0.978} & \textbf{0.993} & \textbf{0.994} \\
\bottomrule
\end{tabular}
\caption{Disfluency detection performance on the PhoDisfluency dataset.}
\label{tab:word_syllable_comparison}
\end{table}
\begin{table}[t]
\centering
\begin{tabular}{lccc}
\toprule
Model & Prec. & Rec. & F1 \\
\midrule
BARTpho          & 0.640 & 0.730 & 0.680 \\
mBART-50         & 0.640 & 0.660 & 0.650 \\
PhoBERT$_{base}$ & 0.670 & 0.780 & 0.720 \\
PhoBERT$_{base}$-v2 & 0.680 & 0.790 & 0.740 \\
PhoBERT$_{large}$ & 0.690 & 0.770 & 0.730 \\
ViDeBERTa$_{base}$ & 0.500 & 0.410 & 0.450 \\
ViT5$_{base}$    & 0.640 & 0.740 & 0.690 \\
XLM-R$_{base}$   & 0.640 & 0.730 & 0.690 \\
XLM-R$_{large}$  & 0.710 & 0.770 & 0.740 \\
\midrule
\textbf{Our Method} & \textbf{0.892} & \textbf{0.893} & \textbf{0.893} \\
\bottomrule
\end{tabular}
\caption{NER performance on the VietMed-NER dataset.}
\label{tab:ner_results}
\end{table}
\begin{table}[t]
\centering
\resizebox{\linewidth}{!}{%
\begin{tabular}{lrrrr}
\toprule
Entity & Prec. & Rec. & F1 & Support \\
\midrule
AGE                & 0.330 & 0.591 & 0.423 & 352   \\
DATETIME           & 0.712 & 0.892 & 0.792 & 968   \\
DIAGNOSTICS        & 0.694 & 0.912 & 0.789 & 570   \\
DISEASESYMTOM      & 0.768 & 0.809 & 0.788 & 1795  \\
DRUGCHEMICAL       & 0.931 & 0.919 & 0.925 & 628   \\
FOODDRINK          & 0.695 & 0.568 & 0.625 & 1205  \\
GENDER             & 0.479 & 0.745 & 0.583 & 1228  \\
LOCATION           & 0.550 & 0.433 & 0.484 & 934   \\
MEDDEVICETECHNIQUE & 0.190 & 0.447 & 0.266 & 235   \\
OCCUPATION         & 0.757 & 0.786 & 0.771 & 1281  \\
ORGAN              & 0.691 & 0.808 & 0.745 & 495   \\
ORGANIZATION       & 0.731 & 0.826 & 0.775 & 679   \\
PERSONALCARE       & 0.636 & 0.762 & 0.693 & 340   \\
PREVENTIVEMED      & 0.964 & 0.935 & 0.949 & 61685 \\
SURGERY            & 0.282 & 0.785 & 0.416 & 219   \\
TRANSPORTATION     & 0.901 & 0.338 & 0.492 & 402   \\
TREATMENT          & 0.752 & 0.773 & 0.762 & 1002  \\
UNITCALIBRATOR     & 0.669 & 0.699 & 0.684 & 2172  \\
\_                 & 0.545 & 0.641 & 0.589 & 573   \\
\midrule
Micro avg          & 0.892 & 0.893 & 0.893 & 76763 \\
Macro avg          & 0.646 & 0.719 & 0.661 & 76763 \\

\bottomrule
\end{tabular}}
\caption{Per-entity results on the VietMed-NER dataset.}
\label{tab:vietmed-ner-results}
\end{table}
Our method achieves strong performance on both the NER and Disfluency Detection tasks, with particularly remarkable gains in Disfluency Detection, where all F1 scores exceed those reported for the fine-tuned PhoBERT\textsubscript{large} in the original study (\cite{dao2022disfluency} for PhoDisfluency; \cite{truong2021covid} for PhoNER-COVID19). It is worth emphasizing that in both benchmark studies, fine-tuned PhoBERT\textsubscript{large} was shown to be a highly competitive baseline, achieving top-tier results. However, our proposed method outperforms it substantially. On the NER task (Table~\ref{tab:ner_comparison}), our model achieves a Micro-F1 score of 0.984, significantly higher than the 0.945 reported for PhoBERT\textsubscript{large} \cite{truong2021covid}. For Disfluency Detection (Table~\ref{tab:word_syllable_comparison}), 
our method attains an exceptional Micro-F1 score of 0.994, approaching perfect accuracy and clearly surpassing the 0.968 score of the fine-tuned PhoBERT\textsubscript{large} reported in \cite{dao2022disfluency}. On the VietMed-NER dataset (Table~\ref{tab:ner_results} and Table~\ref{tab:vietmed-ner-results}), 
our method achieves a substantial improvement with an overall F1 score of 0.892, 
clearly outperforming all baselines reported in \cite{le2024medical}.

The superior performance of our method in token-level classification can be attributed to several key architectural innovations. Our method employs a hybrid architecture that integrates multiple complementary components. Input tokens are first encoded using PhoBERT, a strong Vietnamese pre-trained language model. These contextual embeddings are then linearly projected and passed through a Graph Attention Network (GATConv), which leverages fully-connected graph structures via the \texttt{edge\_index} input. This step enables the model to enrich contextual representations with global token interactions.

The output of the GAT layer is subsequently fed into a Transformer Decoder Layer, which models complex sequential dependencies and facilitates interaction across the entire sequence. Finally, the aggregated representations are passed through a linear classifier to produce output logits for token-level prediction.

\textbf{Compared to BiLSTM-CRF:} While BiLSTM-CRF models sequential dependencies with CRF decoding, it is generally less effective at capturing long-range dependencies. In contrast, our method utilizes transformer-based representations and GAT-enhanced structure modeling, enabling both local and global context understanding.

\textbf{Compared to XLM-R$_{base}$ and XLM-R$_{large}$:} Our method benefits from being grounded in a monolingual model tailored specifically for Vietnamese. While XLM-R offers strong cross-lingual generalization, it may overlook language-specific morphological or syntactic cues that are well captured by PhoBERT. By combining PhoBERT with Vietnamese dependency parses through the GAT layer, our model better exploits the characteristics of the Vietnamese language.

\textbf{Compared to PhoBERT$_{base}$ and PhoBERT$_{large}$:} Our method yields superior results. This improvement is not merely due to further fine-tuning but arises from the hybrid architectural framework that integrates both graph-based and transformer-based modules. The synergy between these components, coupled with joint fine-tuning, allows the model to go beyond the representational power of PhoBERT alone.

In summary, the improvements demonstrated by our model are a direct consequence of the architectural innovations that combine syntactic structure with global context modeling. By jointly fine-tuning all components, we achieve performance gains that surpass strong baselines, including fine-tuned PhoBERT\textsubscript{large}. These findings reinforce the argument that incorporating explicit structural biases, rather than relying solely on backbone models, remains a promising direction for advancing token-level classification in low-resource or linguistically rich languages like Vietnamese.

\subsection{Ablation Study}

To better understand the contribution of each component in our model, 
we conduct an ablation study on the VietMed-NER dataset. 
We compare the following settings: (i) PhoBERT-only baseline, 
(ii) PhoBERT with GAT (without decoder), and (iii) the full model with both GAT and Transformer Decoder (Table~\ref{tab:ablation-vietmed}) . For fair comparison, we report Micro-F1 scores, which are consistent with prior work. The PhoBERT-only baseline scores are taken from \cite{le2024medical}.

\begin{table}[t]
\centering
\resizebox{\linewidth}{!}{%
\begin{tabular}{lccc}
\toprule
\textbf{Model} & \textbf{Prec.} & \textbf{Rec.} & \textbf{Micro-F1} \\
\midrule
PhoBERT-only                          & 0.690 & 0.770 & 0.730 \\
PhoBERT + GAT                         & 0.889 & 0.891 & 0.890 \\
Full (PhoBERT + GAT + Decoder)        & 0.892 & 0.893 & 0.893 \\
\bottomrule
\end{tabular}}
\caption{Ablation study on VietMed-NER.}
\label{tab:ablation-vietmed}
\end{table}

The results show that PhoBERT+GAT consistently improves over the 
PhoBERT-only baseline, demonstrating the effectiveness of graph-based 
token interactions. Although the full model does not significantly 
improve over PhoBERT+GAT in terms of Micro-F1, this behavior is further 
analyzed and explained in Section~5 (Error Analysis), where we show that 
the decoder mainly benefits Macro-F1 and rare entity types.

\section{Error Analysis}
The experimental results reveal a clear performance gap between the model’s performance on the \textbf{VietMed-NER}, \textbf{PhoNER COVID-19}, and \textbf{PhoDisfluency} datasets. 
On \textbf{PhoNER COVID-19} (highest Macro-F1 = 0.958, Micro-F1 = 0.984 see Table~\ref{tab:ner_comparison}) and \textbf{PhoDisfluency} (highest Micro-F1 = 0.994 see Table~\ref{tab:word_syllable_comparison}), the model achieves very high performance, approaching the upper bound. 
This exceptionally high accuracy can be explained by several factors: 

\begin{enumerate}
    \item \textbf{More balanced label distribution and fewer entity types} -- Both datasets contain a limited number of label types (PhoNER COVID-19 focuses mainly on patient-related information and COVID-19 symptoms; PhoDisfluency involves only a few tags related to fluent and disfluent speech), which facilitates faster and more stable learning of entity-specific patterns.  

    \item \textbf{Consistent and domain-specific language patterns} -- Sentences in PhoNER COVID-19 and PhoDisfluency tend to follow consistent syntactic structures and vocabulary specific to their domains (e.g., news articles tagged with COVID-19 from major Vietnamese outlets, conversational transcripts), thereby reducing the contextual variability the model must handle.

    \item \textbf{Absence of rare entities} -- There are no entity types with extremely low frequency, thus avoiding overfitting to small subsets of the data or failing to learn meaningful patterns.  
\end{enumerate}

It is worth noting that the Micro-F1 score of \textbf{0.994 on PhoDisfluency} (see Table~\ref{tab:word_syllable_comparison}) is unusually high, almost reaching perfect accuracy. 
\textbf{This result does not indicate overfitting of the model, but rather reflects the relative simplicity of the dataset: only three label categories, stable discourse patterns, and the absence of rare classes.} 
In other words, the task itself is close to a solved problem under current architectures, and further improvements are naturally bounded by ceiling effects. 
Importantly, this behavior does not generalize across datasets: on VietMed-NER, which contains a larger vocabulary, more entity types, and severe label imbalance, the model’s performance drops substantially (Micro-F1 = 0.893 Table~\ref{tab:vietmed-ner-results}). 
This contrast confirms that the near-perfect score on PhoDisfluency arises from the dataset’s simplicity rather than model overfitting.

In contrast, \textbf{VietMed-NER} (Macro-F1 = 0.661, Micro-F1 = 0.893); see Table~\ref{tab:ner_results} and Table~\ref{tab:vietmed-ner-results})  exhibits noticeably lower performance, particularly for certain entity types such as \textit{MEDDEVICETECHNIQUE} (F1 = 0.266), \textit{SURGERY} (F1 = 0.416), and \textit{AGE} (F1 = 0.423). The main contributing factors include:

\begin{enumerate}
    \item \textbf{Rare entities and imbalanced distribution} -- VietMed-NER contains more than 18 entity types with highly skewed frequency distribution. Certain types occur only a handful of times in the training set, leading to random predictions, susceptibility to noisy samples, and poor generalization. For instance, the label \texttt{TRANSPORTATION} has only 5 training samples with just 2 unique entity mentions, which severely limits the model’s ability to learn representative patterns.

    \item \textbf{Semantic overlap and ambiguous boundaries} — Certain entity types exhibit closely related conceptual scopes or unclear lexical boundaries (e.g., \textit{TREATMENT}, \textit{PREVENTIVEMED}, \textit{SURGERY}), making label assignment prone to confusion.  
\end{enumerate}

In summary, the substantial performance disparity between VietMed-NER and the other two datasets is not inherently due to the model architecture, but rather to the characteristics of the datasets themselves: contextual diversity, the number of entity types, and severe label imbalance.

\paragraph{Impact of Transformer decoder.}
The ablation results show that adding the Transformer decoder consistently improves performance across datasets, though the magnitude of improvement varies with task complexity. On \textbf{VietMed-NER}, a dataset with a large vocabulary and many rare entity types, the decoder yields a substantial Macro-F1 gain (0.571 $\rightarrow$ 0.661), while Micro-F1 increases only slightly (0.890 $\rightarrow$ 0.893). This contrast indicates that the decoder is particularly effective for rare and difficult classes, not just the dominant ones.

By contrast, on medium-scale datasets such as \textbf{PhoNER COVID-19} and \textbf{PhoDisfluency}, the decoder’s effect remains marginal. Macro-F1 improves only from 0.945 $\rightarrow$ 0.958 and 0.980 $\rightarrow$ 0.988, while Micro-F1 rises just from 0.981 $\rightarrow$ 0.984 and 0.990 $\rightarrow$ 0.994 in word level, respectively. These results confirm that the PhoBERT + GAT backbone already achieves near-ceiling Micro-F1 on datasets of moderate complexity, leaving limited room for further gains.

Overall, while Micro-F1 changes are minor across all datasets, the clear Macro-F1 improvements on VietMed-NER highlight the decoder’s role in handling complex domains with unbalanced label distributions. We therefore advocate retaining the Transformer decoder in our full model to robustly address large-scale tasks with richer semantics and many low-frequency entities, whereas PhoBERT + GAT alone is already a strong solution for simpler datasets.

\section{Conclusion}

We introduced a novel architecture for Vietnamese token-level classification, evaluated on both Named Entity Recognition (NER) and Disfluency Detection. Our model consistently outperformed strong baselines such as PhoBERT\textsubscript{large} and XLM-R\textsubscript{large}, achieving state-of-the-art Micro-F1 and Macro-F1 scores. In Disfluency Detection, it nearly reached ceiling performance (Micro-F1 0.993 at the syllable level, 0.994 at the word level see Table~\ref{tab:word_syllable_comparison}) without signs of overfitting.

Despite task-specific fine-tuning, we employed a unified design, showing that the architecture is expressive and adaptable to different token-level problems. To our knowledge, this is the first work combining PhoBERT, Graph Attention, and a Transformer decoder in a single framework for Vietnamese, covering both written and spoken domains. Beyond benchmark gains, the model holds promise for practical applications such as healthcare chatbots, educational dialogue systems, and information extraction. Future directions include multilingual extension and domain adaptation to better support low-resource or specialized settings.

\paragraph{Reproducibility.} 
All code and datasets used in this study are publicly available 
at \href{https://drive.google.com/drive/folders/1sPB4_PdZc2jaV-RXEnsRrzO8MCi6D1mZ?usp=drive_link}{this link}, 
facilitating reproducibility and further research.

\subsection*{Future Directions}

Future research may pursue several promising directions:

\begin{itemize}
  \item \textbf{Multilingual Transfer:} Extending \textbf{TextGraphFuseGAT} to multilingual or zero-shot cross-lingual NER and disfluency detection via shared graph structures.
  \item \textbf{Domain Adaptation:} Leveraging meta-learning or adversarial training for efficient adaptation to new domains (e.g., medical or legal texts) with limited annotation.
  \item \textbf{Multimodal Cues:} Incorporating prosodic (pause, pitch) or visual (gestures) features to enhance disfluency detection in spontaneous speech.
  \item \textbf{Efficient Graphs:} Designing sparse or dynamic graph attention mechanisms to lower computational cost and enable real-time processing.
  \item \textbf{Explainability:} Developing interpretable GAT modules to highlight key token interactions, supporting error analysis and model trust.
  \item \textbf{Applications:} Embedding robust token-level understanding into interactive agents (education, healthcare, dialogue) and deploying in real-world systems such as chatbots or information extraction pipelines.

\end{itemize}

\bibliography{custom}
\end{document}